\pdfoutput=1
\documentclass[11pt,a4paper]{article}

\usepackage[hyperref,acceptedwithA]{tacl2018v2}

\usepackage{times}
\usepackage{latexsym}

\usepackage{xcolor}
\usepackage[normalem]{ulem}
\usepackage{url}
\usepackage{amsmath}
 \usepackage{multirow}

\usepackage{booktabs}
\usepackage{setspace}

\hypersetup{filecolor=darkblue}

\usepackage{microtype}

\title{{G}ender {B}ias in {M}achine {T}ranslation}

\author{Beatrice Savoldi\textsuperscript{1,2}, Marco Gaido\textsuperscript{1,2}, Luisa Bentivogli\textsuperscript{2}, Matteo Negri\textsuperscript{2}, Marco Turchi\textsuperscript{2} \\
  \textsuperscript{1} University of Trento \\
  \textsuperscript{2} Fondazione Bruno Kessler \\
  \texttt{\{bsavoldi,mgaido,bentivo,negri,turchi\}@fbk.eu}}

\date{}

\begin{document}
\maketitle
\begin{abstract}
Machine translation (MT) technology has facilitated our daily tasks by providing accessible shortcuts for gathering, processing and communicating information. However,  it can suffer from biases that harm users and society at large. As a relatively new field of inquiry, studies of gender bias in MT still lack cohesion. This advocates for a unified framework to ease future research. To this end, we: \textit{i)}  critically review current conceptualizations of bias in light of theoretical insights from related disciplines, \textit{ii)} summarize previous analyses aimed at assessing gender bias in MT, \textit{iii)} discuss the mitigating strategies proposed so far, and \textit{iv)} point toward potential directions for future work.
\end{abstract}

\section{Introduction}

Interest in understanding, assessing, and mitigating gender bias is steadily growing within the natural language processing (NLP)  community, with recent studies showing how gender disparities affect language technologies. 
Sometimes, for example, coreference resolution systems fail to recognize women doctors  \cite{zhao2017men,rudinger2018gender}, image captioning models do not detect women sitting next to a computer \cite{hendricks}, and automatic speech recognition works better with male voices \cite{tatman-2017}. 
Despite a prior disregard for such phenomena within research agendas  \cite{cislak2018bias}, it is now widely recognized that NLP tools encode and reflect controversial social asymmetries for many seemingly neutral tasks, machine translation (MT) included.
Admittedly, the problem is not new \cite{frank2004}. A few years ago, \citet{Schiebinger} criticized the phenomenon of ``masculine default" in MT after running one of her interviews through a commercial translation system. In spite of several feminine mentions in the text, she was repeatedly referred to by masculine pronouns.
Gender-related concerns have also been voiced by online MT users, who noticed how commercial systems entrench social gender expectations, e.g., translating engineers as masculine and nurses as feminine \cite{forbes}.

With language technologies entering widespread use and being deployed at a massive scale, their societal impact has raised concern both within \cite{hovy-spruit-2016-social,bender2021dangers} and outside \cite{reuters} the scientific community.
To take stock of the situation, \citet{sun-etal-2019} reviewed NLP studies on the topic. However, their survey is based on monolingual applications, whose underlying assumptions and solutions may not be directly applicable to  languages other than English \cite{zhou2019examining,zhao2020gender,takeshita-etal-2020-existing} and cross-lingual settings.  
Moreover, MT is a multifaceted task, which requires resolving multiple gender-related subtasks at the same time (e.g., coreference resolution,  named entity recognition). Hence, depending on the languages involved and the factors accounted for, gender bias has been conceptualized 
differently across studies. 
To date, gender bias in MT has been tackled by means of a narrow, problem-solving oriented approach. While technical countermeasures are needed, failing to adopt a wider perspective and engage with related literature outside of NLP can be detrimental to the advancement of the field \cite{blodgett-etal-2020-language}. 

In this paper, we intend to put such literature to use for the study of gender bias in MT. We go beyond surveys restricted to monolingual NLP \cite{sun-etal-2019} or more limited in scope \cite{costa-nature,monti2020gender}, and present the first comprehensive review of gender bias in MT. In particular, we \textbf{1)} offer a unified framework that introduces the concepts, sources, and effects of bias in MT, clarified in light of relevant notions on the relation between gender and different languages; \textbf{2)} critically discuss the state of the research by identifying blind spots and key challenges.

\section{Bias statement}
\label{sec:statement}

Bias is a fraught term with partially overlapping, or even competing, definitions \cite{campolo2017ai}. In cognitive science, bias refers to  the possible outcome of
heuristics, i.e., mental shortcuts that can be critical to support prompt reactions \citep{tversky1973availability, tversky1974judgment}. AI research borrowed from such a tradition \cite{rich2019lessons,rahwan2019machine} and conceived bias as the divergence from an ideal or expected value \cite{glymour2019measuring,shah-etal-2020-predictive}, which can occur if models rely on spurious cues and unintended shortcut strategies to predict outputs \citep{schuster-etal-2019-towards,mccoy-etal-2019-right,Geirhos2020}.
Since this can lead to systematic errors and/or adverse social effects, bias investigation is not only a scientific and technical endeavour but also an ethical one, given the growing societal role of NLP applications \cite{bender2018data}. 
As \citet{blodgett-etal-2020-language} recently called out, and has been endorsed in other venues \cite{hardmeier2021write}, analysing bias is an inherently normative process which requires identifying \textit{what} is deemed as harmful behavior, \textit{how}, and to \textit{whom}. Hereby,  we stress a human-centered, sociolinguistically-motivated framing of bias. By drawing on the definition by \citet{friedman1996bias}, we consider as biased an MT model that \textit{systematically} and \textit{unfairly} discriminates against certain individuals or groups in favor of others.
We identify bias  per specific model's behaviors, which are assessed by envisaging their potential risks when the model is deployed \cite{bender2021dangers} and the harms that could ensue \cite{crawford2017trouble}, with people in focus \cite{bendertypology}. Since MT systems are daily employed by millions of individuals, they could impact a wide array of people in different ways.

As a guide, we rely on \citet{crawford2017trouble}, who defines two main categories of harms produced by a biased system: 
\textit{i)} \textbf{Representational} harms (R) -- i.e., detraction from the representation of social groups and their identity, which, in turn, affects attitudes and beliefs; \textit{ii)} \textbf{Allocational} harms (A) -- i.e., a system allocates or withholds opportunities or resources to certain groups.
Considering the so far reported real-world instances of gender bias \cite{Schiebinger,forbes} and those addressed in the MT literature reviewed in this paper, \textbf{(R) }can be further distinguished into \textit{under-representation} and \textit{stereotyping}.

\textit{Under-representation} refers to the reduction of the visibility of certain social groups through language by \textit{i)} producing a disproportionately low representation of women (e.g., most feminine entities in a text are misrepresented as male in translation); or \textit{ii)} not recognizing the existence of non-binary individuals (e.g., when a system does not account for gender neutral forms). For such cases, the misrepresentation occurs in
the language employed to talk ``about'' such groups.\footnote{See 
also the classifications by \citet{Dinan-etal-2020-multi}.}
Also, this harm can imply the reduced visibility of the language used ``by” speakers of such groups by \textit{iii)} failing to reflect their identity and communicative repertoires. 
In these cases, an MT flattens their communication and produces an output that indexes unwanted gender identities and social meanings (e.g. women and non-binary speakers are not referred to by their preferred linguistic expressions of gender).

\textit{Stereotyping} regards the propagation of negative generalizations of a social group, e.g., belittling feminine representation to less prestigious occupations (teacher (Feminine) vs. lecturer (Masculine)), or in association with attractiveness judgments (pretty lecturer (Feminine)). 

Such behaviors are harmful as they can directly affect the self-esteem of members of the target group \cite{Bourguignon:15}. Additionally, they can propagate to indirect stakeholders. For instance, if a system fosters the visibility of the way of speaking of the dominant group, MT users can presume that such a language represents the most appropriate or prestigious variant\footnote{For an analogy on how technology shaped the perception of feminine voices as shrill and immature, see \citet{Tallon}.} – at the expense of other groups and communicative repertoires.
These harms can aggregate, and the ubiquitous embedding of MT in web applications provides us with paradigmatic examples of how the two types of \textbf{(R)} can interplay.
For example, if women or non-binary\footnote{Throughout the paper, we use non-binary as an umbrella term for referring to all gender identities between or outside the masculine/feminine binary categories.} scientists are the subjects of a query, automatically translated pages run the risk of referring to them via masculine-inflected job qualifications.
Such misrepresentations can lead to experience feelings of identity invalidation \cite{zimman2017trans}. Also, users may not be aware of being exposed to MT mistakes due to the deceptively fluent output of a system \cite{martindale-carpuat-2018-fluency}. In the long run, stereotypical assumptions and prejudices (e.g., only men are qualified for high-level positions) will be reinforced \cite{Levesque2011,regner2019committees}.

Regarding \textbf{(A)}, MT services are consumed by the general public and can thus be regarded as resources in their own right. Hence, \textbf{(R)} can directly imply \textbf{(A)} as a performance disparity across users in the \textit{quality of service}, i.e., the overall efficiency of the service.
Accordingly, a woman attempting to translate her biography by relying on an MT system requires additional energy and time to revise wrong  masculine references. If such disparities are not accounted for, the MT field runs the risk of producing systems that prevent certain groups from fully benefiting from such technological resources.

In the following, we operationalize such categories to map studies on gender bias to their motivations and societal implications (Table \ref{tab:assessing} and \ref{tab:mitigating}).

\section{Understanding Bias}

To confront bias in MT, it is vital to reach out to other disciplines that foregrounded how the socio-cultural notions of gender interact with language(s), translation, and implicit biases.
Only then can we discuss the multiple factors that concur to encode and amplify gender inequalities in language technology.
Note that, except for \citet{saunders-etal-2020-neural}, current studies on gender bias in MT have assumed an (often implicit) binary vision of gender. As such, our discussion is largely forced  into this classification. Although we reiterate on bimodal feminine/masculine linguistic forms and social categories, we emphasize that gender encompasses multiple biosocial elements not to be conflated with sex \cite{risman2018gender,fausto2019gender}, and that some individuals do not experience gender, at all, or in binary terms \cite{glen2012measuring}.

\subsection{Gender and Language}

The relation between language and gender is not straightforward.  
First, the linguistic structures  used to refer to the extra-linguistic reality of gender vary across languages ($\S$\ref{subsub:encoding}). Moreover, how gender is assigned and perceived in our verbal practices depends on contextual factors as well as assumptions about social roles, traits, and attributes ($\S$\ref{subsub:connotations}). At last, language is conceived as a tool for articulating and constructing personal identities ($\S$\ref{subsub:use}). 

\subsubsection{Linguistic Encoding of Gender}
\label{subsub:encoding}
Drawing on
\cite{Corbett:91,craig1994classifier,ComrieB1999,Hellinger1,Hellinger2,Hellinger3,Corbett:2013,Gygaxindex4} we hereby  describe the linguistic forms (lexical, pronominal, grammatical) that bear a relation with the extra-linguistic reality of gender. 
Following \citet{stahlberg2007representation}, we identify three language groups: 

\textbf{Genderless languages} (e.g., Finnish, Turkish).  In such languages, the gender-specific repertoire is at its minimum, only expressed for basic lexical pairs, usually kinship or address terms (e.g., in Finnish \textit{sisko}/sister vs. \textit{veli}/brother). 

\textbf{Notional  gender languages}\footnote{Also referred to as \textit{natural}  gender languages. Following \citet{McConnell-Ginet2013}, we prefer notional to avoid terminological overlapping with ``natural'', i.e., biological/anatomical sexual categories. For a wider discussion on the topic, see \citet{NevalainenRaumolin-Brunberg1993,Curzan2003}.}
(e.g., Danish, English). On top of lexical gender (\textit{mom}/\textit{dad}), such languages display a system of pronominal gender (\textit{she/he}, \textit{her/him}). English also hosts some marked derivative nouns (\textit{actor}/\textit{actress}) and compounds (\textit{chairman}/\textit{chairwoman}). 

\textbf{Grammatical gender languages} (e.g., Arabic, Spanish). In these languages, each noun pertains to a class such as masculine, feminine, and neuter (if present).  Although for most inanimate objects gender assignment is only formal,\footnote{E.g., ``moon” is masculine in German, feminine in French.}
for human referents masculine/feminine markings are assigned on a semantic basis. Grammatical gender is defined by a  system of morphosyntactic agreement, where several parts of speech  beside the noun (e.g., verbs, determiners, adjectives) carry gender inflections.

In  light of the above, the English sentence ``\textit{He/She} is a good friend'' has no overt expression of gender in a genderless language like Turkish (``\textit{{O}} iyi bir arkada{ş}''), whereas Spanish spreads several masculine or feminine markings (``\textit{El/\underline{la}} es \textit{un/\underline{a} buen/\underline{a} amig\underline{o}/\underline{a}}''). 
Although general, such macro-categories allow us to highlight typological differences across languages. These are crucial to frame gender issues in both human and machine translation. Also, they exhibit 
to what extent speakers of each group are led to think and communicate via binary distinctions,\footnote{Outside of the Western paradigm, there are cultures whose languages traditionally encode gender outside of the binary  \cite{epple1998coming,murray2003takatapui,hall201410}.}
as well as underline the relative complexity in carving out a space for lexical innovations which encode non-binary gender \cite{hord2016bucking,conrodpronouns}.
In this sense, while English is bringing the singular \textit{they} in common use and developing neo-pronouns \cite{bradley2019singular}, for grammatical gender languages like Spanish neutrality requires the development of neo-morphemes (``\textit{El\underline{le}} es \textit{un\underline{e} buen\underline{e} amig\underline{ue}'')}.

\subsubsection{Social Gender Connotations} 
\label{subsub:connotations}

To understand gender bias, we have to grasp not only the structure of different languages, but also how linguistic expressions are connoted, deployed, and perceived \cite{Hellinger&Motschenbacher:2015}.  In grammatical gender languages, feminine forms are often subject to a so-called semantic derogation \cite{Schulz1975}, e.g., in French, \textit{couturier} (fashion designer) vs. \textit{couturière} (seamstress). English is no exception (e.g., \textit{governor}/\textit{governess}). 

Moreover, bias can lurk underneath seemingly neutral forms. Such is the case of epicene (i.e., gender neutral) nouns where gender is not grammatically marked. Here, gender assignment is linked to (typically binary) social gender, i.e.,  ``the socially imposed dichotomy of masculine and feminine role and character traits'' \cite{Kramarae1985}. 
As an illustration, Danish speakers tend to pronominalize \textit{dommer} (judge) with \textit{han} (he) when referring to the whole occupational category \cite{Gomard1995,nissen2013aspects}. 
Social gender assignment varies across time and space \cite{lyons1977semantics,Romaine:1999,Cameron2003} and regards stereotypical assumptions about what is typical or appropriate for men and women. Such assumptions impact our perceptions \cite{hamilton1988using,Gygaxetal2008,Kreineretal2008} and influence our behavior -- e.g., leading individuals to identify with and fulfill stereotypical expectations \cite{hannover2002auswirkungen,sczesny2019agency} -- and verbal communication, e.g.,  women are often misquoted in the academic community \cite{krawczyk2017all}.

Translation studies highlight how social gender assignment influences translation choices \cite{jakobson1959linguistic,Chamberlain1988,ComrieB1999,di2020grammatical}. Primarily, the problem arises from typological differences across languages and their gender systems. Nonetheless, socio-cultural factors also
influence how translators deal with such differences. 
Consider the character of the cook in Daphne du Maurier's ``Rebecca'', whose gender is never explicitly stated in the whole book. In the lack of any available information, translators of five grammatical gender languages represented the character as either a man or a woman \cite{Wandruszka:1969,nissen2013aspects}.
Although extreme, this case can illustrate the situation of uncertainty faced by MT: the  mapping of one-to-many forms in gender prediction. But, as discussed in $\S$\ref{subsec:stereotyping}, mistranslations occur when contextual gender information is available as well.

\subsubsection{Gender and Language Use} 
\label{subsub:use}

Language use varies between demographic groups and reflects their backgrounds, personalities, and social identities \cite{labov1972sociolinguistic,Trudgill:2000,PennebakerStone2003}.
In this light, the study of gender and language variation has received much attention in socio- and corpus linguistics \cite{HolmesMeyerhoff:2003,EckertMcConnell-Ginet2003}.
Research conducted in speech and text analysis highlighted several gender differences, which are exhibited at the phonological and lexical-syntactic  level. For example, women rely more on  hedging strategies (``it seems that''), purpose clauses (``in order to''), first-person pronouns, and prosodic exclamations \cite{Mulac2001,Mondorf2002,brownlow2003gender}.
Although some correspondences between gender and linguistic features hold across cultures and languages \cite{smith2003gendered,johannsen2015cross}, it should be kept in mind that they are far from universal\footnote{It has been largely debated whether gender-related differences are inherently biological or cultural and social products \cite{Mulac2001}. Currently, the idea that they
depend on biological reasons is largely rejected \cite{hyde2005gender} in favor of a socio-cultural or performative perspective \cite{butler2011gender}.}
and should not be intended in a stereotyped and oversimplified manner \cite{Bergvall1996,nguyen2016computational,koolen2017these}. 

Drawing on gender-related features proved useful to build demographically informed NLP tools \cite{garimella2019women} and personalized MT models \cite{mirkin2015motivating,bawden:hal-01353860,rabinovich2017personalized}.
However, using personal gender as a variable requires a prior understanding of which categories may be salient, and a critical reflection on how gender is intended and ascribed \cite{larson-2017gender}. Otherwise, if we assume that the only relevant (sexual) categories are ``male'' and ``female'', our models will inevitably fulfill such a reductionist expectation \cite{bamman2014gender}.

\subsection{Gender Bias in MT}
\label{subsec:biasMT}

To date, an overview of how several factors may contribute to gender bias in MT does not exist. We identify and clarify concurring problematic causes, accounting for the context in which systems are developed and used ($\S$\ref{sec:statement}). 
To this aim, we rely on the three overarching categories of bias described by~\citet{friedman1996bias}, which foreground different sources that can lead to machine bias.
These are: pre-existing bias -- rooted in our institutions, practices and attitudes  ($\S$\ref{subsub:pre}),  technical bias -- due to technical constraints and decisions ($\S$\ref{subsub:tech}), and emergent bias  -- arising from the interaction between systems and users ($\S$\ref{subsub:emergent}). We consider such categories as placed along a continuum, rather than being discrete.

\subsubsection{Pre-existing Bias}
\label{subsub:pre}

MT models are known to reflect gender disparities present in the data. However, reflections on such generally invoked disparities are often overlooked. 
Treating data as an abstract, monolithic entity \cite{gitelman2013raw} -- or relying on ``overly broad/overloaded terms like \textit{training data bias}''\footnote{See \citep{venturebeat,machineindifferent} for a discussion on how such narrative can be counterproductive for tackling bias.} \cite{suresh2019framework} -- do not encourage reasoning on the many factors of which data are the product. First and foremost, the historical, socio-cultural context in which they are generated.

A starting point to tackle these issues is the Europarl corpus \cite{koehn2005europarl}, where only 30\% of sentences are uttered by women \cite{vanmassenhove-etal-2018}. Such an imbalance is a direct window into the glass ceiling that has hampered women’s access to parliamentary positions. This case exemplifies how  data might be ``tainted with historical bias”, mirroring  an ``unequal ground truth” \cite{hacker2018teaching}. 
However, other gender variables are harder to spot and quantify.

Empirical linguistics research pointed out that subtle gender asymmetries are rooted in languages’ use and structure. For instance, an important aspect regards how women are referred to. Femaleness is often explicitly invoked when there is no textual need to do so, even in languages that do not require overt gender marking. 
A case in point regards Turkish, which differentiates \textit{cocuk} (child) and \textit{kiz cocugu} (female child) \cite{braun2000geschlecht}. Similarly, in a corpus search, \citet{romaine2001corpus} found 155 explicit female markings for \textit{doctor} (female, woman or lady doctor), compared to only 14 \textit{male doctor}. Feminist language critique provided extensive analysis of such a phenomenon by highlighting how referents in discourse are considered men by default unless explicitly stated  \cite{silveira1980generic,hamilton1991masculine}. 
Finally, prescriptive top-down guidelines limit the linguistic visibility of gender diversity, e.g., the Real Academia de la Lengua Española recently discarded the official use of non-binary innovations and claimed the functionality of masculine generics \cite{mundo,artemisRAE}.

By stressing such issues, we are not condoning the reproduction of pre-existing bias in MT. Rather, the above-mentioned concerns are the starting point to account for when dealing with gender bias.

\subsubsection{Technical Bias}
\label{subsub:tech}

Technical bias comprises aspects related to data creation, models design, training and testing procedures.
If present in training and testing samples, asymmetries in the semantics of language use and gender distribution are respectively learnt by MT systems and rewarded in their evaluation. However, as just discussed, biased representations are not merely quantitative, but also qualitative. Accordingly, straightforward procedures -- e.g., balancing the number of speakers in existing datasets -- do not ensure a fairer representation of gender in MT outputs.
Since datasets are a crucial source of bias, it is also crucial to advocate for a careful data curation \cite{mehrabi2019survey,paullada-2020-data, denton, bender2021dangers}, guided by pragmatically- and socially-informed analyses \cite{hitti-etal-2019-proposed, sap2020social,devinney-etal-2020-semi} and annotation practices \cite{gaido-etal-2020-breeding}.

Overall, while data can mirror gender inequalities and offer adverse shortcut learning opportunities, it is ``quite clear that data alone rarely constrain a model sufficiently'' \cite{Geirhos2020} nor  explain the fact that models \textit{overamplify} \cite{shah-etal-2020-predictive} such inequalities in their outputs.
Focusing on models' components, \citet{costajussa2020gender} demonstrate that architectural choices in multilingual MT impact the systems’ behavior: shared encoder-decoders retain less gender information in the source embeddings and less diversion in the attention than language-specific encoder-decoders  \cite{escolano2020multilingual}, thus disfavoring the generation of feminine forms.
While discussing the loss and decay of certain words in translation,  \citet{vanmassenhove-etal-2019, eva2020} attest to the existence of an algorithmic bias that leads under-represented forms in the training data -- as it may be the case for feminine references -- to further decrease in the MT output. Specifically, \citet{roberts2020decoding} prove that beam search -- unlike sampling -- is skewed toward the generation of more frequent (masculine) pronouns, as it leads models to an extreme operating point that exhibits zero variability.

Thus, efforts towards understating and mitigating gender bias should also account for the model front. To date, this remains largely unexplored.

\subsubsection{Emergent Bias}
\label{subsub:emergent}

Emergent bias may arise when a system is used in a different context than the one it was designed for, e.g., when it is applied to another demographic group. From car crash dummies to clinical trials, we have evidence of how not accounting for gender differences brings to the creation of male-grounded products with dire consequences  \cite{liu2016women,criado2019invisible}, such as higher death and injury risks in vehicle crash and less effective medical treatments for women.
Similarly, unbeknownst to their creators, MT systems that are not intentionally envisioned for a diverse range of users will not generalize for the feminine segment of the population. Hence, in the interaction with an MT system, a woman will likely be misgendered or not have her linguistic style preserved  \cite{hovy-etal-2020-sound}.
Other conditions of users/system mismatch may be the result of changing societal knowledge and values. A case in point regards Google Translate’s historical decision to adjust its system for instances of gender ambiguity. Since its launch twenty years ago, Google had provided only one translation for single-word gender-ambiguous queries (e.g., \textit{professor} translated in Italian with the masculine \textit{professore}).
In a community increasingly conscious of the power of language to hardwire stereotypical beliefs and women’s invisibility \cite{lindqvist2019reducing,beukeboom2019stereotypes}, the bias exhibited by the system was confronted with a new sensitivity. The service’s decision \cite{google2018} to provide a double feminine/masculine output (\textit{professor}$\rightarrow$\textit{professoressa}$\mid$\textit{professore}) stems from current demands for gender-inclusive resolutions.
For the recognition of non-binary groups \cite{richards2016non}, we invite studies on how such modeling could be integrated with neutral strategies ($\S$\ref{sec:future}). 

\section{Assessing  Bias}

First accounts on gender bias in MT date back to \citet{frank2004}. Their manual analysis pointed out how English-German MT suffers from a dearth of linguistic competence, as it shows severe difficulties in recovering syntactic and semantic information to correctly produce  gender agreement.

Similar inquiries were conducted on other target grammatical gender languages for several commercial MT systems \cite{abu2017errors,monti2017,rescigno-etal-2020-case}. While these studies focused on contrastive phenomena, \citet{Schiebinger}\footnote{See also Schiebinger's project \textit{Gendered Innovations}: \url{http://genderedinnovations.stanford.edu/case-studies/nlp.html}} went beyond linguistic insights, calling for a deeper understanding of gender bias. 
Her article on Google Translate's ``masculine default'' behavior emphasized how such a phenomenon is related to the larger issue of gender inequalities, also perpetuated by socio-technical artifacts \cite{selbst2019fairness}. 
All in all, these qualitative analyses demonstrated that gender problems encompass all three MT paradigms (neural, statistical, and rule-based), preparing the ground for quantitative work. 

To attest the existence and scale of gender bias across several languages, dedicated benchmarks, evaluations, and experiments have been designed. We first discuss large scale analyses aimed at assessing gender bias in MT, grouped according to two main conceptualizations: \textit{i)} works focusing on the weight of prejudices and stereotypes in MT ($\S$\ref{subsec:stereotyping}); \textit{ii)} studies assessing whether gender is properly  preserved in translation ($\S$\ref{subsec:quality}).
In accordance with the human-centered approach embraced in this survey, in Table~\ref{tab:assessing} we map each work to the harms (see $\S$\ref{sec:statement}) ensuing from the biased behaviors they assess. Finally, we review existing benchmarks for comparing MT performance across genders ($\S$\ref{subsec:benchmarks}).

\subsection{MT and Gender Stereotypes}
\label{subsec:stereotyping}

In MT, we record prior studies concerned with pronoun translation and coreference resolution  across typologically different languages accounting for both animate and inanimate referents \cite{hardmeier2010modelling,le2010aiding,guillou2012improving}. For the specific analysis on gender bias, instead, such tasks are exclusively studied in relation to human entities.

\citet{Prates2018AssessingGB} and \citet{cho-etal-2019measuring} design a similar setting to assess gender bias. \citet{Prates2018AssessingGB} investigate pronoun translation from 12 genderless languages into English. 
Retrieving $\sim$1,000  job positions from the U.S. Bureau of Labor Statistics, they build simple constructions like the Hungarian ``\textit{ő} egy \textit{mérnök}" (``\textit{he/she} is an \textit{engineer}''). Following the same template, \citet{cho-etal-2019measuring} extend the analysis to Korean-English including both occupations and sentiment words (e.g., \textit{kind}). 
As their samples are ambiguous by design, the observed predictions of he/she pronouns should be random, yet they show a strong masculine skew.\footnote{\citet{cho-etal-2019measuring} highlight that a higher frequency of feminine references in the MT output does not necessarily imply a bias reduction. Rather, it may reflect gender stereotypes, as for \textit{hairdresser} that is skewed toward feminine. This observation points to the tension between frequency count, suitable for testing under-representation, and qualitative-oriented analysis on bias conceptualized in terms of stereotyping.}

To further analyze the under-representation of \textit{she} pronouns, \citet{Prates2018AssessingGB} focus on 22 macro-categories of occupation areas and compare the proportion of pronoun predictions against the real-world proportion of men and women employed in such sectors.
In this way, they find that MT not only yields a masculine default, but it also underestimates feminine frequency at a greater rate than occupation data alone suggest.
Such an analysis starts by acknowledging pre-existing bias (see $\S$\ref{subsub:pre}) -- e.g., low rates of women in STEM -- to attest the existence of machine bias, and defines it as the exacerbation of actual gender disparities.

Going beyond word lists and simple synthetic constructions, \citet{gonen2020automatically} inspect the translation into Russian, Spanish, German, and French of natural yet ambiguous English sentences. Their analysis on the ratio and type of generated masculine/feminine job titles consistently exhibits social asymmetries for target grammatical gender languages (e.g., \textit{lecturer} masculine vs. \textit{teacher} feminine).
Finally, \citet{stanovsky-etal-2019} assess that MT is skewed to the point of actually ignoring explicit feminine gender information in source English sentences. For instance, MT systems yield a wrong masculine translation of the job title \textit{baker}, although it is referred to by the pronoun \textit{she}. Beside the overlook of overt gender mentions, the model's reliance on unintended (and irrelevant) cues for gender assignment is further confirmed by the fact that adding a socially connoted -- but formally epicene -- adjective (the \textit{pretty} baker) pushes models toward feminine inflections in translation.

We observe that the propagation of stereotypes is a widely researched form of gender asymmetries in MT, one that so far has been largely narrowed down to occupational stereotyping. After all, occupational stereotyping has been studied by different disciplines \cite{greenwald1998measuring} attested across cultures \cite{lewis2020gender}, and it can be easily detected in MT across multiple language directions with consistent results. Current research should not neglect other stereotyping dynamics, as in the case of \citet{stanovsky-etal-2019} and \citet{cho-etal-2019measuring}, who include associations to physical characteristics or psychological traits. Also, the intrinsically contextual nature of societal expectations advocates for the study of culture-specific dimensions of bias. 
Finally, we signal that the BERT-based perturbation method by \citet{webster-etal-2019-gendered} identifies other bias-susceptible nouns that tend to be assigned to a specific gender (e.g., \textit{fighter} as masculine). 
As \citet{Blodget-thesis} underscores, however, ``the existence of these undesirable correlations is not sufficient to identify them as normatively undesirable''. It should thus be investigated whether such statistical preferences can cause harms, e.g., by checking if they map to existing harmful associations or quality of service disparities.

\subsection{MT and Gender Preservation}
\label{subsec:quality}

\citet{vanmassenhove-etal-2018} and \citet{hovy-etal-2020-sound} investigate whether speakers' gender\footnote{Note that these studies distinguish speakers into female/male. As discussed in $\S$\ref{subsub:use}, we invite a reflection on the appropriateness and use of such categories.}
is properly reflected in MT. 
This line of research is preceded by findings on gender personalization of statistical MT \cite{mirkin2015motivating,bawden:hal-01353860,rabinovich2017personalized}, which claim that gender ``signals'' are weakened in translation. 

\citet{hovy-etal-2020-sound} conjecture the existence of age and gender stylistic bias due to models' under-exposure to the writings of women and younger segments of the population. To test this hypothesis, they automatically translate a corpus of online reviews with available metadata about users \cite{trustpilot}. Then, they compare such demographic information with the prediction of age and gender classifiers run on the MT output. Results indicate that different commercial  MT models systematically make authors 
``sound'' older and male. 
Their study thus concerns the under-representation of the language used ``by'' certain speakers and how it is perceived \cite{Blodget-thesis}.
However, the authors do not inspect which linguistic choices MT overproduces, nor which stylistic features may characterize different socio-demographic groups.
 
Still starting from the assumption that demographic factors influence language use, \citet{vanmassenhove-etal-2018} probe MT's ability to preserve speaker's gender translating from English into ten languages. To this aim, they develop gender-informed MT models (see $\S$~\ref{subsec:debiasing}), whose outputs are compared with those obtained by their baseline counterparts. Tested on a  set for spoken language translation \cite{koehn2005europarl}, their enhanced models show consistent gains in terms of overall  quality when translating into grammatical gender languages, where speaker's references are often marked. For instance, the French translation of ``I'm  \textit{happy}'' is either ``Je suis \textit{heureuse}`` or ``Je suis \textit{hereux}'' for a female/male speaker respectively. 
Through a focused cross-gender analysis -- carried out by splitting their English-French test set into 1st person male  vs. female data -- they assess that the largest margin of  improvement for their gender-informed approach concerns sentences uttered by women, since the results of their baseline disclose a quality of service disparity in favor of male speakers. 
Besides morphological agreement, they also attribute such 
improvement to the fact that their enhanced model produces gendered preferences in other word choices. 
For instance, it opts for \textit{think} rather than \textit{believe}, which is in concordance with corpus studies claiming a tendency for women to use less assertive speech \cite{newman2008gender}. 
Note that the authors rely on manual analysis to ascribe performance differences to gender-related features. In fact, global evaluations on generic test sets alone  are inadequate to pointedly measure gender bias.

\begin{table*}[t]
\centering
\setlength{\tabcolsep}{3pt}
\footnotesize
\begin{tabular}{|l|l|l|l|}
\hline
\textbf{Study} & \textbf{Benchmark} & \textbf{Gender} & \textbf{Harms} \\
\hline
 \cite{Prates2018AssessingGB} & Synthetic, U.S. Bureau of Labor Statistics & b & R: under-rep, stereotyping \\ \hline
 \cite{cho-etal-2019measuring} & Synthetic equity evaluation corpus (EEC) & b & R: under-rep, stereotyping \\ \hline
  \cite{gonen2020automatically} & BERT-based perturbations on natural sentences & b & R: under-rep, stereotyping \\ \hline
 \cite{stanovsky-etal-2019} & WinoMT & b & R: under-rep, stereotyping \\ \hline
 \cite{vanmassenhove-etal-2018} & Europarl (generic) & b & A: quality \\ \hline
 \cite{hovy-etal-2020-sound} & Trustpilot (reviews with gender and age) & b & R: under-rep \\ \hline
 
\end{tabular}
  \caption{
  \footnotesize{
  For each 
  \textbf{Study}, the Table shows on which \textbf{Benchmark} 
  gender bias is assessed,
   how \textbf{Gender} is intended (here only in binary (b) terms). 
  Finally, we indicate which (R)epresentational -- \textit{under-representation} and \textit{stereotyping}
  -- or (A)llocational \textbf{Harm} -- as reduced \textit{quality} of service -- is addressed in the study}.}

  \label{tab:assessing}
\end{table*}

\subsection{Existing Benchmarks}
\label{subsec:benchmarks}

MT outputs are typically evaluated against reference translations employing standard metrics such as BLEU~\cite{papineni-etal-2002bleu} or TER~\cite{Snover:06}. This procedure poses two challenges. 
First, these metrics provide coarse-grained scores for translation quality, as they treat all errors equally and are rather insensitive to  specific linguistic phenomena \cite{sennrich2017grammatical}. Second,  generic test sets containing the same gender imbalance  present in the training data can reward biased predictions. 
Hereby, we describe the publicly available MT \textit{Gender Bias Evaluation Testsets} (GBETs) \cite{sun-etal-2019},  i.e., benchmarks designed to probe gender bias by isolating the impact of gender from other factors that may affect systems' performance. Note that different benchmarks and metrics respond to different conceptualizations of bias \cite{barocas2018fairness}. Common to them all in MT, however, is that biased behaviors are formalized by using some variants of averaged performance\footnote{This is a value-laden option \cite{abeba}, and not the only possible one \cite{mitchell2020diversity}. For a broader discussion on measurement and bias we refer the reader also to \cite{jacobs2019measurement,blodgetttutorial}.} disparities across gender groups, comparing the accuracy of gender predictions on an equal number of masculine, feminine, and neutral references.

\citet{escude-font-costa-jussa-2019equalizing} developed the bilingual English-Spanish \textbf{Occupations test set.} It consists of 1,000 sentences equally distributed across genders. The phrasal 
structure envisioned for their sentences is ``I’ve known \textit {\{her$\mid$him$\mid<$proper noun$>$\}} for a long time, my friend works as \textit \textit{\{a$\mid$an\} $<$occupation$>$}''. The evaluation focuses on the translation of the noun \textit{friend} into Spanish (\textit{amig\underline{o}/\underline{a}}).
Since gender information is present in the source context and sentences are the same for both masculine/feminine participants, an MT system exhibits gender bias if it disregards relevant context and cannot provide the correct translation of \textit{friend} at the same rate across genders.

\citet{stanovsky-etal-2019} created \textbf{WinoMT} by concatenating two existing English GBETs for coreference resolution \cite{rudinger2018gender,zhao-etal-2018}. The corpus consists of 3,888 Winogradesque sentences presenting two human entities defined by their role and a subsequent pronoun that needs to be correctly resolved to one of the entities (e.g., ``The \textit{lawyer} yelled at the \textit{hairdresser} because \textit{he} did a bad job''). For each sentence, there are two variants with either \textit{he} or \textit{she} pronouns, so as to cast the referred annotated entity (\textit{hairdresser}) into a proto- or anti-stereotypical gender role. By translating WinoMT into grammatical gender languages, one can thus measure systems' ability to resolve the anaphoric relation and pick the correct feminine/masculine inflection for the occupational noun. 
On top of quantifying under-representation as the difference between the total amount of translated feminine and masculine references, the subdivision of the corpus into proto- and anti-stereotypical sets also allows verifying if MT predictions correlate with occupational stereotyping.

Finally, \citet{saunders-etal-2020-neural} enriched the original version of WinoMT in two different ways. First, they included a third gender-neutral case based on the singular \textit{they} pronoun, thus paving the way to account for non-binary referents. Second, they labeled the entity in the sentence which is not coreferent with the pronoun (\textit{lawyer}). The latter annotation is used to verify the shortcomings of some mitigating approaches as discussed in $\S$\ref{sec:mitigating}.

The above-mentioned corpora are known as \textit{challenge sets}, consisting of sentences created \textit{ad hoc} for diagnostic purposes. In this way, they  can be used to quantify bias related to stereotyping and under-representation in a sound environment. However, since they consist of a limited variety of synthetic gender-related phenomena, they hardly address the variety of challenges posed by real-world language and are relatively easy to overfit. As recognized by \citet{rudinger2018gender} ``they may demonstrate the presence of gender bias in a system, but not prove its absence''.

\textbf{The Arabic Parallel Gender Corpus} \cite{habash-etal-2019} includes an English-Arabic  test set\footnote{Overall, the corpus comprises  over  12,000 annotated sentences and  200,000 synthetic sentences.} 
retrieved from OpenSubtitles natural language data \cite{lison-tiedemann-2016opensubtitles2016}. Each of the 2,448 sentences in the set exhibits a  first person singular reference to the speaker (e.g., ``I'm \textit{rich}''). Among them, $\sim$200 English sentences require gender agreement to be assigned in translation. These were translated into Arabic in both gender forms, obtaining a quantitatively and qualitatively equal amount of sentence pairs with annotated masculine/feminine references. This natural corpus thus allows for cross-gender evaluations on MT production of correct speaker's gender agreement.

\textbf{MuST-SHE} \cite{bentivogli-etal-2020-gender} is a natural benchmark for three language pairs (English-French/Italian/Spanish). Built on TED talks data \cite{Cattoni2020mustc-v2}, for each language pair it comprises  $\sim$1,000  (\textit{audio}, \textit{transcript}, \textit{translation}) triplets, thus allowing evaluation for both MT and speech translation (ST). Its samples are balanced between masculine and feminine phenomena, and incorporate two types of constructions: \textit{i)} sentences referring to the speaker (e.g., ``\textit{I} was \textit{born} in Mumbai''), and  \textit{ii)} sentences that present contextual information to disambiguate gender (e.g., ``My \textit{mum} was \textit{born} in Mumbai''). 
Since every gender-marked word in the target language is annotated in the corpus, MuST-SHE grants the advantage of complementing BLEU- and accuracy-based evaluations on gender translation for a great variety of phenomena.

Unlike challenge sets, natural corpora quantify whether MT yields reduced feminine representation in authentic conditions and whether the quality of service varies across speakers of different genders. However, as they treat all gender-marked words equally, it is not possible to identify if the model is propagating stereotypical representations.

All in all, we stress that each test set and metric is only a proxy for framing a phenomenon or an ability (e.g., anaphora resolution), and an approximation of what we truly intend to gauge. Thus, as we discuss in $\S$\ref{sec:future}, advances in MT should account for the observation of gender bias in real-world conditions to avoid that achieving high scores on a mathematically formalized esteem could lead to a false sense of security. Still, benchmarks remain valuable  tools to monitor models' behavior. As such, we remark that evaluation procedures ought to cover both models’ general performance and gender-related issues. This is crucial to establish the 
capabilities and limits of mitigating  strategies.

\section{Mitigating Bias}
\label{sec:mitigating}

To attenuate gender bias in MT, different strategies dealing with input data, learning algorithms, and model outputs have been proposed. 
As attested by \citet{abeba}, since advancements are oftentimes exclusively reported in terms of values internal to the machine learning field (e.g efficiency, performance), it is not clear how such strategies are meeting societal needs by reducing MT-related harms.
In order to conciliate technical perspectives with the intended social purpose, in Table \ref{tab:mitigating} we map each mitigating approach to the harms (see $\S$\ref{sec:statement}) they are meant to alleviate, as well as to the benchmark their effectiveness is evaluated against. Complementarily, we hereby describe each approach by means of two categories: model debiasing ($\S$\ref{subsec:debiasing}) and debiasing through external components ($\S$\ref{subsec:specific}).

\begin{table*}[t]
\centering
\setlength{\tabcolsep}{3pt}
\footnotesize
\begin{tabular}{|p{2.5cm}|p{3.8cm}|p{4cm}|l|p{3.5cm}|}
\hline
\textbf{Approach} & \textbf{Authors} & \textbf{Benchmark} & 
 \textbf{Gender} & \textbf{Harms} \\
\hline
\multirow{2}{*}{\begin{tabular}[c]{@{}l@{}}Gender tagging \\ (sentence-level)\end{tabular}} & \citeauthor{vanmassenhove-etal-2019} & Europarl (generic) & b & R: under-rep, A: quality \\ \cline{2-5}  & \citeauthor{Elaraby2018GenderAS} & Open subtitles (generic) & 
 b &
 R: under-rep, A: quality \\ \hline
\multirow{2}{*}{\begin{tabular}[c]{@{}l@{}}Gender tagging \\ (word-level)\end{tabular}} & \citeauthor{saunders-etal-2020-neural} & expanded WinoMT  
&  
nb
& R: under-rep, stereotyping \\ \cline{2-5} 
 & \citeauthor{stafanovics-etal-2020-mitigating} & WinoMT 
 & b & R: under-rep, stereotyping \\ \hline
Adding context & \citeauthor{basta-etal-2020-towards} & WinoMT 
& b & R: under-rep, stereotyping \\ \hline
Word-embeddings & \citeauthor{escude-font-costa-jussa-2019equalizing} & Occupation test set 
& b & R: under-rep \\ \hline
Fine-tuning & \citeauthor{costa-jussa-de-jorge-2020-fine} & WinoMT & b & R: under-rep, stereotyping \\ \hline
Black-box injection & \citeauthor{moryossef-etal-2019} & Open subtitles (selected sample)
& b & R: under-rep, A: quality \\ \hline
Lattice-rescoring & \citeauthor{saunders-byrne-2020-reducing} & WinoMT 
& b & R: under-rep, steretoyping \\ \hline
Re-inflection & \citeauthor{habash-etal-2019,alhafni-etal-2020-gender} & Arabic  Parallel Gender Corpus
& b & R: under-rep, A: quality \\ \hline
\end{tabular}
  \caption{
  \footnotesize{
  For each  \textbf{Approach} and related \textbf{Authors}, the Table shows on which \textbf{Benchmark} it is tested, if  \textbf{Gender} is intended in binary terms (b), or including non-binary (nb) identities. Finally, we indicate which (R)epresentational -- \textit{under-representation} and \textit{stereotyping} -- or (A)llocational \textbf{Harm} -- as reduced \textit{quality} of service -- the approach attempts to mitigate.}}
  \label{tab:mitigating}
\end{table*}

\subsection{Model Debiasing}
\label{subsec:debiasing}

This line of work focuses on mitigating gender bias through architectural changes of general-purpose MT models or via dedicated training procedures.

\textbf{Gender tagging.} 
To improve the generation of speaker's referential markings, \citet{vanmassenhove-etal-2018} prepend a gender tag (M or F) to each source sentence, both at training and inference time. As their model is able to leverage this additional information, the approach proves useful to handle morphological  agreement when translating from English into French. However, this solution requires additional metadata regarding the speakers'
gender that might not always be feasible to acquire. 
Automatic annotation of speakers' gender (e.g., based on 
first names) is not advisable, as it runs the risk of introducing additional bias by making unlicensed assumptions about one's identity.

\citet{Elaraby2018GenderAS} 
bypass this risk by defining a comprehensive set of cross-lingual gender agreement rules based on POS tagging. In this way, they identify speakers' and listeners' gender references in an English-Arabic parallel corpus, which is consequently labeled and used for training. 
The idea, originally developed for spoken language translation in a two-way conversational setting, can be adapted for other languages and scenarios by creating new dedicated rules. 
However, in realistic deployment conditions where reference translations are not available, gender information still has to be externally supplied as metadata at inference time.

\citet{stafanovics-etal-2020-mitigating} and \citet{saunders-etal-2020-neural}
explore the use of word-level gender tags. While \citet{stafanovics-etal-2020-mitigating} just report a gender translation improvement, \citet{saunders-etal-2020-neural} rely on the expanded version of WinoMT to identify a problem concerning gender tagging: it introduces noise if applied to sentences with
references to multiple participants, as it pushes their translation toward the same gender.
\citet{saunders-etal-2020-neural} also include a first non-binary exploration of neutral translation by exploiting an artificial dataset, where neutral tags are added and gendered inflections are replaced by placeholders. The results are however inconclusive, most likely due to the small size  and synthetic nature of their dataset.

\textbf{Adding context.}
Without further information needed for training or inference, \citet{basta-etal-2020-towards} adopt a generic approach and concatenate each sentence with its preceding one. By providing more context, they attest a slight  improvement in gender translations requiring anaphorical coreference to be solved in English-Spanish. This finding motivates exploration at the document level, but it should be validated with manual \cite{castilho-etal-2020-context} and interpretability analyses since the added context can be beneficial for gender-unrelated reasons, such as acting as a regularization factor \cite{kim-etal-2019-document}.

\textbf{Debiased word embeddings.} 
The two above-mentioned mitigations share the same intent: supply the model with additional gender knowledge. Instead, \citet{escude-font-costa-jussa-2019equalizing} leverage pre-trained word embeddings, which are debiased by using the hard-debiasing method proposed by \citet{bolukbasi2016man} or the GN-GloVe algorithm \cite{zhao-etal-2018-embeddings}. 
These methods respectively remove gender associations or isolate them from the representations of English gender-neutral words. \citet{escude-font-costa-jussa-2019equalizing} employ such embeddings on the decoder side, the encoder side, and both sides of an English-Spanish model. The best results are obtained by leveraging GN-GloVe embeddings on both encoder and decoder sides, increasing BLEU scores and gender accuracy. 
The authors generically apply debiasing methods developed for English also to their target language. However, being Spanish a grammatical gender language, other language-specific approaches should be considered to preserve the quality of the original embeddings \cite{zhou2019examining, zhao2020gender}.
We also stress that it is debated whether depriving systems of some knowledge and ``blind'' their perceptions is the right path toward fairer language models \citep{dwork2012fairness,caliskan2017semantics, gonen-goldberg-2019,nissim2020fair}. 
Also, \citet{goldfarb2020intrinsic} find that there is no reliable correlation between intrinsic evaluations of bias in word-embeddings and cascaded effects on MT models' biased behavior.

\textbf{Balanced fine-tuning.} \citet{costa-jussa-de-jorge-2020-fine} rely on Gebiotoolkit~\cite{costa2020gebiotoolkit} to build gender-balanced datasets (i.e., featuring an equal amount of masculine/feminine references) based on Wikipedia biographies. 
By fine-tuning their models on such natural and more even data, the generation of feminine forms is overall improved. However, the approach is not as effective for gender translation on the anti-stereotypical WinoMT set.
As discussed in $\S$\ref{subsub:tech}, they employ a straightforward method that aims to increase the amount of feminine Wikipedia pages in their training data. However, such coverage increase does not mitigate stereotyping harms, as it does not account for the qualitative different ways in which men and women are portrayed \cite{wagner2015s}.

\subsection{Debiasing through External Components}
\label{subsec:specific} 
Instead of directly debiasing the MT model, these mitigating strategies intervene in the inference phase with external dedicated components. Such approaches do not imply retraining, but introduce the additional cost of maintaining separate modules and handling their integration with the MT model.

\textbf{Black-box injection.}  
\citet{moryossef-etal-2019} attempt to control the production of feminine references
to the speaker and numeral inflections (plural or singular) for the listener(s) in an English-Hebrew spoken language setting. To this aim, they rely on a short construction, such as ``\textit{she} said to \textit{them}'', which is prepended to the source sentence and then removed from the MT output. 
Their approach is simple, it can handle two types of information (gender and number) for multiple entities (speaker and listener), and improves systems' ability to generate feminine target forms. However, as in the case of \citet{vanmassenhove-etal-2018} and \citet{Elaraby2018GenderAS}, it requires metadata about speakers and listeners.

\textbf{Lattice re-scoring.}
\citet{saunders-byrne-2020-reducing} propose to post-process the MT output with a lattice re-scoring module. This module exploits a transducer to create a lattice by  mapping gender marked words in the MT output to all their possible inflectional variants.
Developed for German, Spanish, and Hebrew, all the sentences corresponding to the paths in the lattice are re-scored with another model, which has been gender-debiased but at the cost of lower generic translation quality. Then, the sentence with the highest probability is picked as the final output. When tested on WinoMT, such an approach leads to an increase in the accuracy of gender forms selection. 
Note that the gender-debiased system is created by fine-tuning the model on an \textit{ad hoc} built tiny set containing a balanced amount of masculine/feminine forms. Such an approach, also known as \textit{counterfactual data augmentation} \cite{lu2019gender}, 
requires to create identical pairs of sentences differing only in terms of gender references. In fact, \citet{saunders-byrne-2020-reducing} compile English sentences following this schema:  ``The $<$profession$>$ finished \textit{$<$his$\mid$her$>$} work''. Then, the sentences are automatically translated and manually checked. In this way, they obtain gender-balanced parallel corpus. Thus,  to implement their method for other language pairs, 
the generation of new data is necessary. 
For the fine-tuning set, the effort required is limited as the goal is to alleviate stereotypes 
by focusing on a pre-defined occupational lexicon. However, data augmentation is 
very demanding for complex sentences that represent a rich variety of gender agreement phenomena\footnote{\citet{zmigrod2019counterfactual} proposed an automatic approach for  augmenting data into morphologically-rich languages, but it is only viable for simple constructions with one single entity.} such as 
those occurring in natural language scenarios.

\textbf{Gender re-inflection.}
\citet{habash-etal-2019} and \citet{alhafni-etal-2020-gender} confront the problem of speaker's gender agreement in Arabic with a post-processing component that re-inflects 1st person references into masculine/feminine forms.
In \citet{alhafni-etal-2020-gender}, the preferred gender of the speaker and the translated Arabic sentence are fed to the component, which re-inflects the sentence in the desired form. In \citet{habash-etal-2019} the component can be: \textit{i}) a two-step system that first identifies the gender of 1st person references in an MT output, and then re-inflects them in the opposite form; \textit{ii}) a single-step system that always produces both forms from an MT output. 
Their method does not necessarily require speakers' gender information: if metadata are supplied, the MT output is re-inflected accordingly; differently, both  feminine/masculine inflections  are offered (leaving to the user the choice of the appropriate one).
The implementation of the re-inflection component was made possible by the Arabic Parallel Gender Corpus (see $\S$\ref{subsec:benchmarks}), which demanded an expensive work of manual data creation. However, such corpus  grants research on English-Arabic the benefits of a wealth of gender-informed natural language data that have been curated  to avoid hetero-centrist interpretations and  preconceptions (e.g., proper names and speakers of sentences like ``that’s my wife'' are flagged as gender-ambiguous).
Along the same line, Google Translate also delivers two outputs for short gender-ambiguous queries \cite{google2020}. Among languages with grammatical gender, the service is 
currently available only for English-Spanish.

In light of the above, we remark that there is no conclusive state-of-the-art method for mitigating bias. The discussed interventions in MT tend to respond to specific aspects of the problem with modular solutions, but if and how they can be integrated within the same MT system remains unexplored. As we have discussed through the survey, the umbrella term ``gender bias’’ refers to a wide array of undesirable phenomena. Thus, it is unlikely that a one-size-fits-all solution will be able tackle problems that differ from one another, as they depend on e.g., how bias is conceptualized, the language combinations, the kinds of corpora used. As a result, we believe that generalization and scalability should not be the only criteria against which mitigating strategies are valued. Conversely, we should make room for openly context-aware interventions.
Finally, gender bias in MT is a socio-technical problem. We thus highlight that engineering interventions alone are not a panacea \cite{changetutorial} and should be integrated with long-term multidisciplinary commitment and practices \cite{d2020data, gebruoxford} necessary to address bias in our community, hence in its artifacts, too.

\section{Conclusion and Key Challenges} 
\label{sec:future}

As studies confronting gender bias in MT are rapidly emerging, in this paper we presented them within a unified framework to critically overview current conceptualizations and approaches to the problem. Since gender bias is a multifaceted and  interdisciplinary issue, in our discussion we integrated knowledge from related disciplines, which can be instrumental to guide future research and make it thrive.  
We conclude by suggesting several directions that can help this field going forward.

\textbf{Model de-biasing.}
Neural networks rely on easy-to-learn shortcuts or ``cheap tricks'' \cite{levesque-2014-on}, as picking up on spurious correlations offered by training data can be easier for machines than learning to actually solve a specific task. 
What is ``easy to learn'' for a model depends on the \textit{inductive bias} \cite{sinz-et-al-2019-engineering,Geirhos2020} resulting from architectural choices, training data and learning rules.
We think that explainability techniques~\citep{belinkov2020linguistic}  
represent a useful tool to identify spurious cues (features) exploited by the model during inference. Discerning them can provide the research community with guidance on how to improve models' generalization by working on data, architectures, loss functions and optimizations. For instance, data responsible for spurious features (e.g., stereotypical correlations) might be recognized and their weight at training time might be lowered
\cite{karimi-mahabadi-etal-2020-end}.
Besides, state-of-the-art architectural choices and algorithms in MT have mostly been studied in terms of overall translation quality without specific analyses regarding gender translation. For instance, current systems segment text into subword units with statistical methods that can break the morphological structure of words, thus losing relevant semantic and syntactic information in morphologically-rich languages \cite{niehues-etal-2016pre,ataman:17}.
Several languages show complex feminine forms, typically derivative and created by adding a suffix to the masculine form, such as \textit{Lehrer/Lehrer\underline{in}} (de), \textit{studente/studente\underline{ssa}} (it). 
It would be relevant to investigate whether, compared to other segmentation techniques, statistical approaches disadvantage (rarer and more complex) feminine  forms. The MT community should not overlook  focused hypotheses of such kind, as they can deepen our comprehension of the gender bias conundrum.

\textbf{Non-textual modalities.} 
Gender bias for non-textual automatic translations (e.g., audiovisual) has been largely neglected.  In this sense, ST represents a small niche \citep{costa2020evaluating}. 
For the translation of speaker-related gender phenomena, \citet{bentivogli-etal-2020-gender} prove that direct ST systems exploit speaker’s vocal characteristics as a gender cue to improve feminine translation. However, as addressed by \citet{gaido-etal-2020-breeding}, relying on physical gender cues (e.g., pitch) for such task implies reductionist gender classifications \citep{zimmantransgender} making systems potentially harmful for a diverse range of users.
Similarly, although image-guided translation has been claimed useful for gender translation since it relies on visual inputs for disambiguation \citep{frank2018assessing, ive-etal-2019-distilling}, it could bend toward stereotypical assumptions about appearance. 
Further research should explore such directions to identify potential challenges and risks, by drawing on bias in image captioning \cite{van2019pragmatic} and consolidated studies from the fields of automatic gender recognition and human-computer interaction (HCI)  \cite{hamidi,keyes,may2019}.

\textbf{Beyond Dichotomies.}
Besides a few  notable exceptions for English NLP tasks \cite{manziniblack,cao-daume-iii-2020-toward,sun2021they} and one in MT \cite{saunders-etal-2020-neural}, the discussion around gender bias has been reduced to the binary masculine/feminine dichotomy. Although research in this direction is currently hampered by the absence of data, we invite considering inclusive solutions and exploring nuanced dimensions of gender. 
Starting from language practices, Indirect Non-binary Language (INL) overcomes gender specifications (e.g., using \textit{service, humankind} rather than \textit{waiter/waitress} or \textit{mankind}).\footnote{INL suggestions have also been recently implemented within Microsoft text editors \cite{microsoft}.}
Whilst more challenging, INL can be achieved also for grammatical gender languages \cite{motschenbacher2014grammatical,lindqvist2019reducing}, and it is endorsed for official EU documents \cite{Euguideline}.
Accordingly, MT models could be brought to avoid binary forms and move toward gender-unspecified solutions, e.g., adversarial networks including a discriminator that classifies speaker's linguistic expression of gender (masculine or feminine) could be employed to ``neutralize'' speaker-related forms  \cite{li-etal-2018-towards,delobelle2020ethical}.
Conversely, Direct Non-binary Language (DNL) aims at increasing the visibility of non-binary individuals via neologisms and neomorphemes \cite{bradley2019singular,papadopoulos2019morphological,knisely2020franccais}. With DNL starting to circulate \cite{shroy2016innovations,vocalfries,linternaartemis}, the community is presented with the opportunity to promote the creation of inclusive data.

Finally, as already highlighted in legal and social science theory, discrimination can arise from the intersection of multiple identity categories (e.g., race and gender) \cite{crenshaw1989demarginalizing} which are not additive and cannot always be detected in isolation \cite{interHCI}. Following the MT work by \citet{hovy-etal-2020-sound}, as well as other intersectional analyses from NLP  \cite{herbelot-etal-2012,jiang-fellbaum-2020-interdependencies} and AI-related fields \cite{buolamwini2018gender}, future studies may account for the interaction of gender attributes with other sociodemographic classes.

\textbf{Human-in-the-loop.} 
Research on gender bias in MT is still restricted to lab tests. As such, unlike other studies that rely on participatory design \cite{TURNER2015136,cercas-curry-etal-2020-conversational,unmet}, the advancement of the field is not measured with people's experience in focus or in relation to specific deployment contexts. However, these are fundamental considerations to guide the field forward and, as HCI studies show \cite{HCI}, to propel the creation of gender-inclusive technology.
In particular, representational harms are intrinsically difficult to estimate and available benchmarks only provide a rough idea of their extent. This advocates for focused studies\footnote{To the best of our knowledge, the \textit{Gender-Inclusive Language Models Survey} is the first project of this kind that includes MT. At time of writing it is available at: \href{url}{https://
docs.google.com/forms/d/e/1FAIpQLSfKenp4RKtDhKA0W
LqPflGSBV2VdBA9h3F8MwqRex\_4kiCf9Q/viewform}} on their individual or aggregate effects 
in everyday life.
Also, we invite the whole development process to be paired with bias-aware research methodology \cite{havens-etal-2020-situated} and HCI approaches \cite{stumpf2020gender}, which can help to operationalize sensitive attributes like gender \cite{os}. 
Finally, MT is not only built for people, but also by people. Thus, it  is  vital  to  reflect on the implicit biases and backgrounds of the people involved in MT pipelines at all stages and how they could be reflected in the model. This means starting from bottom-level countermeasures, engaging with translators \cite{gendertraining,lessinger2020challenges}, annotators \cite{waseem-2016-racist,task/annotator}, considering everyone’s subjective positionality and, crucially, also the lack of diversity within technology teams \cite{schluter-2018glass,waseem2020disembodied}.  

\section*{Acknowledgments}
We would like to thank the anonymous reviewers and the TACL Action Editors. Their insightful comments helped us improve on the current version of the paper.

\bibliography{tacl2021_final}
\bibliographystyle{acl_natbib}

\end{document}